\documentclass[conference]{IEEEtran}
\IEEEoverridecommandlockouts

\makeatletter
\let\NAT@parse\undefined
\makeatother
\usepackage[numbers]{natbib}

\usepackage{sty/irap_acronyms}
\usepackage{sty/irap_math}
\usepackage{sty/irap_SIunits}
\usepackage{sty/irap_misc}

\usepackage{amsmath,amssymb,amsfonts}
\usepackage{mathtools}
\usepackage{algorithmic}
\usepackage{graphicx}
\usepackage{textcomp}
\usepackage{xcolor}
\usepackage{caption}

\usepackage{subfigure}

\usepackage{soul,color}
\usepackage{verbatim} 
\usepackage{multirow}

\usepackage{lipsum}
\usepackage{xcolor}

\def\BibTeX{{\rm B\kern-.05em{\sc i\kern-.025em b}\kern-.08em
    T\kern-.1667em\lower.7ex\hbox{E}\kern-.125emX}}

\begin{document}

\title{SC-LiDAR-SLAM: a Front-end Agnostic \\ Versatile LiDAR SLAM System \\
\thanks{† This research was supported by a grant(21TSRD-B151228-03) from Urban Declining Area Regenerative Capacity-Enhancing Technology Research Program funded by Ministry of Land, Infrastructure and Transport of Korean government.}
}

\author{
    
\IEEEauthorblockN{1\textsuperscript{st} Giseop Kim}
\IEEEauthorblockA{\textit{Dept. of Civil and Envtl. Eng.} \\
\textit{KAIST}\\
Daejeon, South Korea \\
paulgkim@kaist.ac.kr}

\and

\IEEEauthorblockN{2\textsuperscript{rd} Seungsang Yun}
\IEEEauthorblockA{\textit{Robotics Program} \\
\textit{KAIST}\\
Daejeon, South Korea \\
seungsang@kaist.ac.kr}

\and 

\IEEEauthorblockN{3\textsuperscript{nd} Jeongyun Kim}
\IEEEauthorblockA{\textit{Dept. of Civil and Envtl. Eng.} \\
\textit{KAIST}\\
Daejeon, South Korea \\
jungyun0609@kaist.ac.kr}

\and

\IEEEauthorblockN{4\textsuperscript{th} Ayoung Kim${}^{*}$}
\IEEEauthorblockA{\textit{Dept. of Mechanical Eng.} \\
\textit{Seoul National University}\\
Seoul, South Korea \\
ayoungk@snu.ac.kr}

} 

\maketitle

\begin{abstract}

Accurate 3D point cloud map generation is a core task for various robot missions or even for data-driven urban analysis. To do so, \ac{LiDAR} sensor-based \ac{SLAM} technology have been elaborated. To compose a full \ac{SLAM} system, many odometry and place recognition methods have independently been proposed in academia. However, they have hardly been integrated or too tightly combined so that exchanging (upgrading) either single odometry or place recognition module is very effort demanding. Recently, the performance of each module has been improved a lot, so it is necessary to build a SLAM system that can effectively integrate them and easily replace them with the latest one. In this paper, we release such a front-end agnostic LiDAR SLAM system, named SC-LiDAR-SLAM. We built a complete SLAM system by designing it modular, and successfully integrating it with Scan Context\texttt{++} and diverse existing open-source LiDAR odometry methods to generate an accurate point cloud map.

\end{abstract}
    
    
\section{Introduction}
\label{sec:intro}

Point cloud \cite{rusu20113d} is a very primitive yet direct representation that could accurately render 3D data such as objects or surrounding structural environments \cite{qi2017pointnet}. Large-scale reconstruction of a 3D real-world environment represented in a point cloud, in particular, is required for a wide range of applications from autonomous robot navigation \cite{krusi2017driving} to data-driven urban analysis \cite{KIM201974}.

\begin{figure}[!t]
  \centering
  \includegraphics[width=0.99\columnwidth, trim = 0 0 0 0, clip]{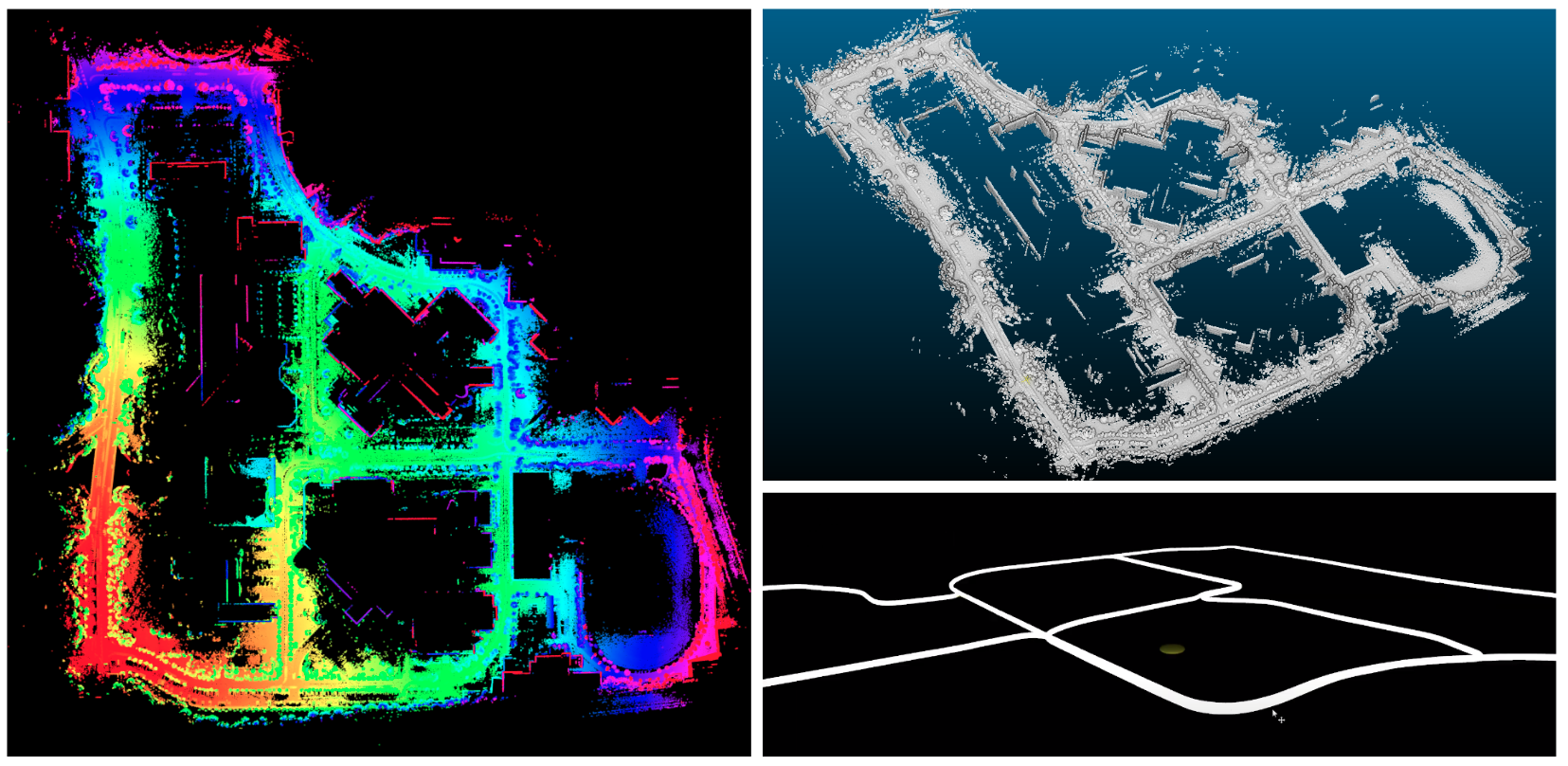}
  \caption{
    An example result of the proposed LiDAR SLAM system on \texttt{KAIST} sequence of \texttt{MulRan} dataset. We can see the point cloud map (left: top-view, right-up: side-view) is qualitatively precise and its optimized trajectory (right-bottom) is well-aligned without drifts particularly on the z-axis.
  }
  \label{fig:kaist}
\end{figure}

Recent advances in \ac{LiDAR} sensor-based robotic mapping technology have made it possible to create a dense and precise 3D point cloud map \cite{kim2020mulran} (e.g., \figref{fig:kaist}). Its general pipeline, typically known as \ac{SLAM} \cite{cadena2016past}, is constituted of three independent modules: odometry \cite{shan2018lego, lin2019loam_livox, cho2020unsupervised}, place recognition \cite{he2016m2dp, kim2018scan, Chen2019OverlapNetLC}, and pose-graph optimization \cite{kaess2008isam, kaess2012isam2, sunderhauf2012switchable}. Seamless integration of the aforementioned three modules is particularly critical for map accuracy because standalone odometry is prone to drifts in motion estimation. Many improved LiDAR odometry approaches have recently been proposed \cite{shan2020lio, li2021towards, xu2021fast, lv2021clins} to overcome this motion estimation error, but it remains unavoidable.

However, there are few open-source LiDAR SLAM projects that have integrated the entire modules, especially the lack of global place recognition (i.e., recognizing previously visited places without an initial information, also called kidnapped robot problem). Thus, current publicly available LiDAR SLAM methods are still vulnerable to accumulated drifts and noisy map generation (e.g., the before case in \figref{fig:riverside3}). To resolve the accumulated errors, a set of long-term data associations made by place recognition is vital.

\begin{figure*}[!t]
  \centering
  \includegraphics[width=0.86\textwidth, trim = 0 0 0 0, clip]{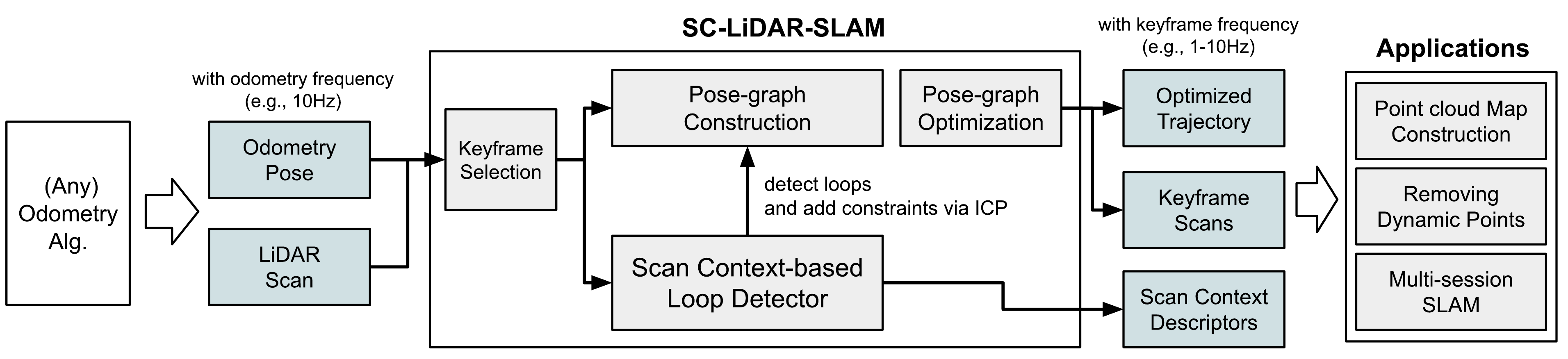}
  \caption{
    The overall pipeline of the proposed robotic mapping system, named SC-LiDAR SLAM.
  }
  \label{fig:scpgo}
\end{figure*}


In this paper, we propose a front-end agnostic LiDAR SLAM system named SC-LiDAR-SLAM and release it as an open-source publicly available. SC-LiDAR-SLAM is \textit{front-end agnostic}, which means it can build a precise 3D point cloud map using any LiDAR odometry method an user prefers with absolutely no additional efforts. Based on our modular implementation design, Scan Context \cite{kim2018scan, gskim-2021-tro} is chosen for global place recognition and internally incorporated with the generic input data from any given odometry method. Finally, SC-LiDAR-SLAM is bundled with useful utilities such as data saver, which stores SLAM results that could be plugged into other robot missions.   

Our contributions are threefolds.

\begin{itemize}
    \item We open a LiDAR SLAM system called SC-LiDAR-SLAM to the public. To the best of our knowledge, this is the first front-end agnostic LiDAR SLAM project made available to the academic community.
    \item To verify the versatility of SC-LiDAR-SLAM, we demonstrate many integrated results in various LiDAR odometry methods and multiple environments.
    \item We provide multiple tools (e.g., to save file-based outputs) to support a set of plug-and-play-based other robot applications such as dynamic point removal or multi-session map building that take our SLAM outputs without additional data conversion efforts, in addition to our robust and versatile LiDAR SLAM system.    
\end{itemize}
  
\section{Backgrounds}
\label{sec:related}

A simple yet complete form of a modern SLAM system can be defined as \cite{cadena2016past}: 
\begin{align}
\begin{split}
        \text{SLAM} =& \ \text{Odometry} \\
                    +& \ \text{Place Recognition}\\ 
                    +& \ \text{Pose-graph Optimization} \ . 
\end{split}
\label{eq:slamdef}
\end{align}
We present here existing works of each component based on the aforementioned notion of SLAM. 

First, odometry or ego-motion estimation is a backbone of a SLAM system, which estimates consecutive relative motions while a robot moves in space. Raw (or downsampled) point cloud registration famously known as \ac{ICP} \cite{segal2009generalized, pomerleau2013comparing} or feature-based methods represented by LOAM \cite{zhang2014loam, zhang2017low} have been contributed to many recent advanced LiDAR odometry methods showing more accurate results \cite{behley2018efficient, shan2018lego, shan2020lio, li2021towards, xu2021fast}. Despite these gains in accuracy, it is still laborious to avoid the accumulation of the motion estimation errors. This accumulated error causes inconsistency between point cloud maps when a robot revisits the same place. Thus, we need to minimize the error by linking temporally distant places and adding a constraint between them. This procedure is called loop closing and it is carried out using place recognition.

Driven by this need, diverse LiDAR sensor-based place recognition methods have been proposed \cite{he2016m2dp,dube2017segmatch, angelina2018pointnetvlad,kim2018scan, Chen2019OverlapNetLC}. Recently, in our previous work, we proposed a rotation and lateral invariant point cloud descriptor named Scan Context\texttt{++} \cite{gskim-2021-tro}. It summarizes a single LiDAR scan into two perpendicular axes which are invariant to and variant to robot motions (e.g., lane change or reversed heading) in urban road environments. Scan Context\texttt{++} takes highest $z$ value of points to capture an ego-centric context of a point cloud this encoding function is inspired by the ubran analysis concept Isovist \cite{KIM201974}

Despite the progress in each odometry and place recognition field, however, their improvements have commonly been isolated and rarely been integrated to construct a state-of-the-art SLAM system following the definition of \equref{eq:slamdef}. Alternatively, in the case of the exising LiDAR SLAM system (e.g., \cite{dube2017segmatch,dube2020segmap}), odometry and place recognition modules are tightly entangled, making it difficult to replace or update one improved module at a low cost. To overcome these limitations, in this paper, we propose a front-end (especially odometry) agnostic modular LiDAR SLAM system, SC-LiDAR-SLAM. The details of our framework are followed in the next section and in \figref{fig:scpgo}.








\section{SC-LiDAR-SLAM}
\label{sec:method}

\subsection{Overview}
\label{sec:method1}

The overall pipeline is depicted in \figref{fig:scpgo}. First, an initial pose-graph is being constructed using a stream of inputs (i.e., relative motion estimation and raw LiDAR scan) from any existing LiDAR odometry method and any LiDAR sensor. Second, thanks to the powerful revisit detection performance of Scan Context, a set of constraints between temporally apart locations could be added to the pose-graph. Finally, an efficient pose-graph optimization \cite{kaess2012isam2} is followed, and the optimized poses with of places' data are saved. Those optimized poses and the placewise data such as LiDAR scan acquired a a place or a place descriptor (i.e., Scan Context Descriptor) would be effective materials for next robot missions (e.g., \secref{sec:result3} and \secref{sec:result3}).

\subsection{Pose-graph Construction}
\label{sec:method2}

Based on the SLAM definition \equref{eq:slamdef}, the robot poses are managed by a pose-graph \cite{grisetti2010tutorial, kim2010multiple, carlone2015initialization, mendes2016icp, walcott2012dynamic} without optimizing landmarks. Before optimizing the pose variables in a pose-graph, pose-graph construction is preceded. The pose-graph construction can be classified in two types.

\subsubsection{Short-term Data Association} 

Short-term data association is a process of initializing a new pose variable and connecting it with its consecutive previous variable. This sequential constraint is conducted by any LiDAR odometry methods. However, this set of relative motion estimations is prone to have errors, which need to be resolved by long-term data association, called loop closing.

\subsubsection{Long-term Data Association}

Long-term data association entails creating a virtual observation constraint between two poses $X_j$ and $X_k$ that are separated in time. The necessity of this association is recognized by the global place recognition module, here, Scan Context\texttt{++} \cite{gskim-2021-tro}. Scan Context\texttt{++} summarizes a raw point cloud of a keyframe into a fixed-size of descriptor for fast retrieval. It is rotation and lateral invariant, making it suitable for robotic mapping in urban areas with changing revisit conditions.

After the previously visited place $j$ is detected, a physical 6D contraint $z_{jk}$ (i.e., the relative translation and orientation between the two poses) is made via \ac{ICP} by registering a query LiDAR scan at $X_k$ to a submap point cloud surrounding the matched pose $X_j$. Finally, this constraint is added to the pose-graph. It is required to minimize the accumulated odometry errors over a traverse. This long-term data association is generally called \textit{loop closing} because it is made when a robot detects previously visited places when it forms its trajectory as a loop, and the robot "closes" the errors to make a complete and well-aligned loop.

\subsubsection{Pose-graph Optimization}

Pose-graph optimization step finds the best solution of the set of poses based on the short-term or the long-term constraints in a pose-graph. This optimization is a MAP (maximum a posteriori) inference \cite{dellaert2017factor} that minimizes a sum of nonlinear least-square errors. For example, the cost function can be written as 
\begin{align}
\begin{split}
        X^{\text{MAP}} = \operatorname*{argmin}_{X} &\sum_{i} \|h(X_i, X_{i+1}) - z_{i} \|^2_{\Sigma_{o_i}}  \\
    + &\sum_{(j, k)} \| s_{jk} \left( h(X_{j}, X_{k}) - z_{jk} \right) \|^2_{\Sigma_{l_{jk}}} \ ,
    \label{eq:pgo}
\end{split}
\end{align}
where $i$ is an index of a pose. $j$ and $k$ is a pair of indexes of a detected loop by Scan Context\texttt{++}. $s_{jk}$ is a scaling factor \cite{agarwal2013robust} for a loop factor between the pose $j$ and $k$. It is used only for loop factors for minimizing the effect of spurious loops (i.e., falsely recognized places), which may trigger catastrophic map collapses. Commonly used M-estimators such as Cauchy or Huber kernels \cite{zhang1997parameter, chebrolu2021adaptive} can be also adopted for this purpose instead of the dynamic scaling. $h(\cdot)$ is a relative transformation between the two argument poses. $z_i$ and $z_{jk}$ are measured by odometry and \ac{ICP} after the place recognition. $\Sigma_o$ and $\Sigma_l$ are covariance matrices for odometry and loop constraints, which determine the uncertainties of each measurement. 

To solve the above problem, a traditional Gauss-Newton style iterative update \cite{kummerle2011g,grisetti2020least} or iSAM-based \cite{mkaess-isam, kaess2012isam2} solvers are used.

\subsection{Open-sources and Utility}
\label{sec:method3}

\begin{figure}[!t]
  \centering
  \includegraphics[width=0.99\columnwidth, trim = 0 0 110 0, clip]{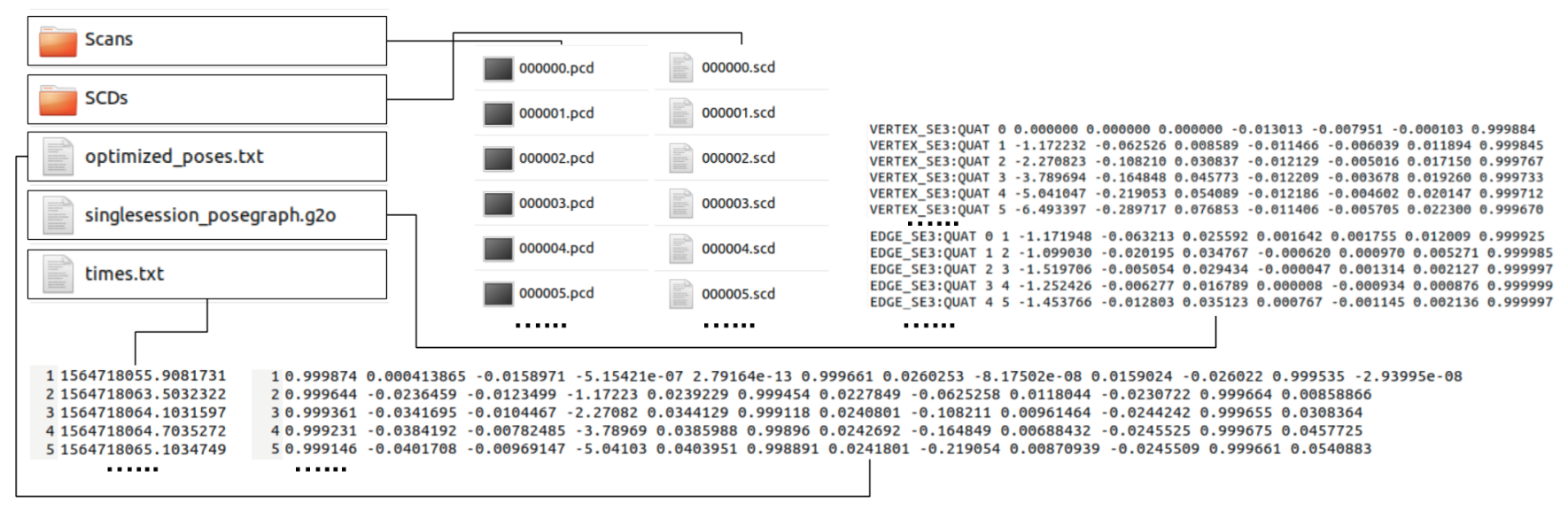}
  \caption{
    A data strucutre of SC-LiDAR-SLAM's saved outputs are visualized. These data (i.e., LiDAR point cloud scans, the optimized trajectory, and place descriptors) are stored place-wise.
  }
  \label{fig:saver}
\end{figure}

\begin{figure}[!t]
  \centering
  \includegraphics[width=0.99\columnwidth, trim = 0 0 0 0, clip]{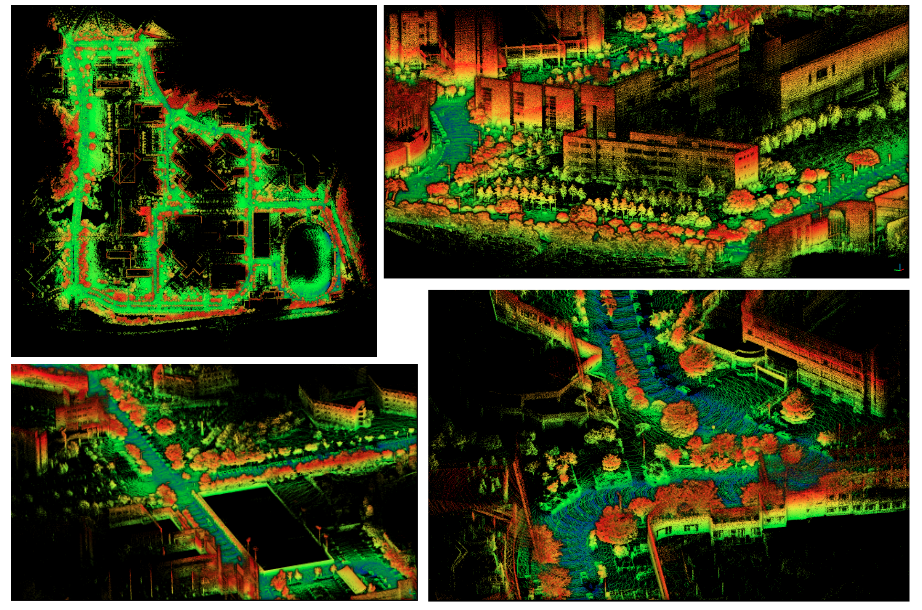}
  \caption{
    Using the optimized poses, the SC-LiDAR-SLAM map making tool created a 3D point cloud map (\texttt{KAIST} sequence of \texttt{MulRan} dataset).
  }
  \label{fig:kaistmap}
\end{figure}

We support to save SLAM outputs in compatible formats to our other robot application softwares such as dynamic point removal \cite{gskim-2020-iros} or multi-session map merging \cite{kim2021lt}. The stored file structure is visualized in \figref{fig:saver}. Also, we prepare a set of tools such as 3D map maker as in \figref{fig:kaistmap} using the optimized poses and LiDAR scans saved as files.  

\section{Result}
\label{sec:result}

\begin{figure*}[!t]
  \centering
  \includegraphics[width=0.86\textwidth, trim = 0 0 0 0, clip]{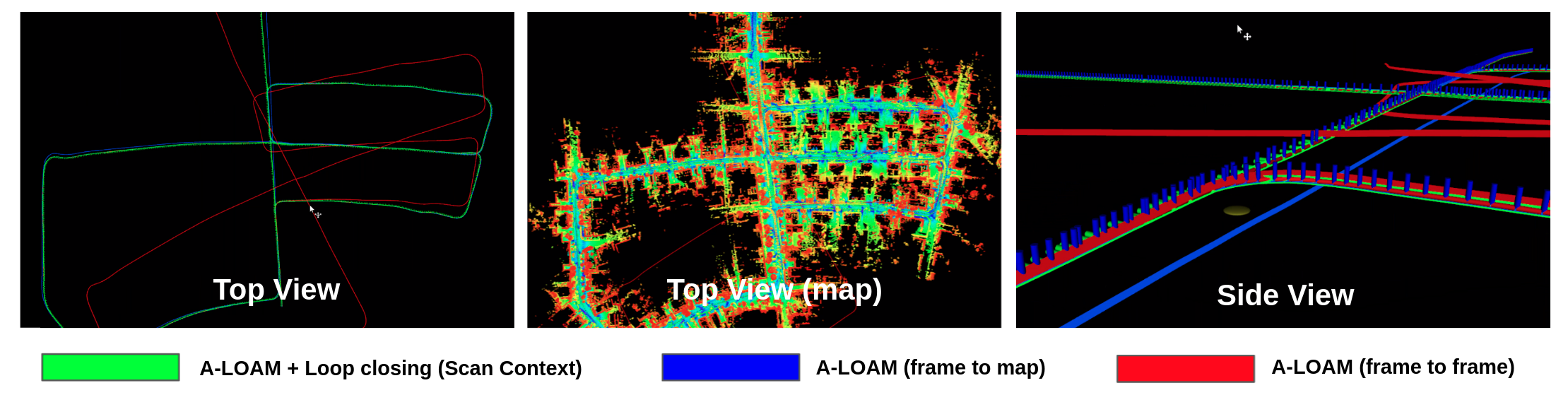}
  \caption{
    Result visualizations of SC-LiDAR-SLAM integrated with A-LOAM. The odometry only method without being equipped with global place recognition (i.e., A-LOAM, the blue lines) inevitably contained accumulated drifts, which is particularly noticeable on the z-axis (see the right plot, a zoomed side-view), from errors of relative motion estimation at every step. 
  }
  \label{fig:kitti05}
\end{figure*}


\begin{figure}[!t]
  \centering
  \subfigure[ \texttt{Riverside} sequence's wide urban roads ]{%
    \includegraphics[width=0.98\columnwidth]{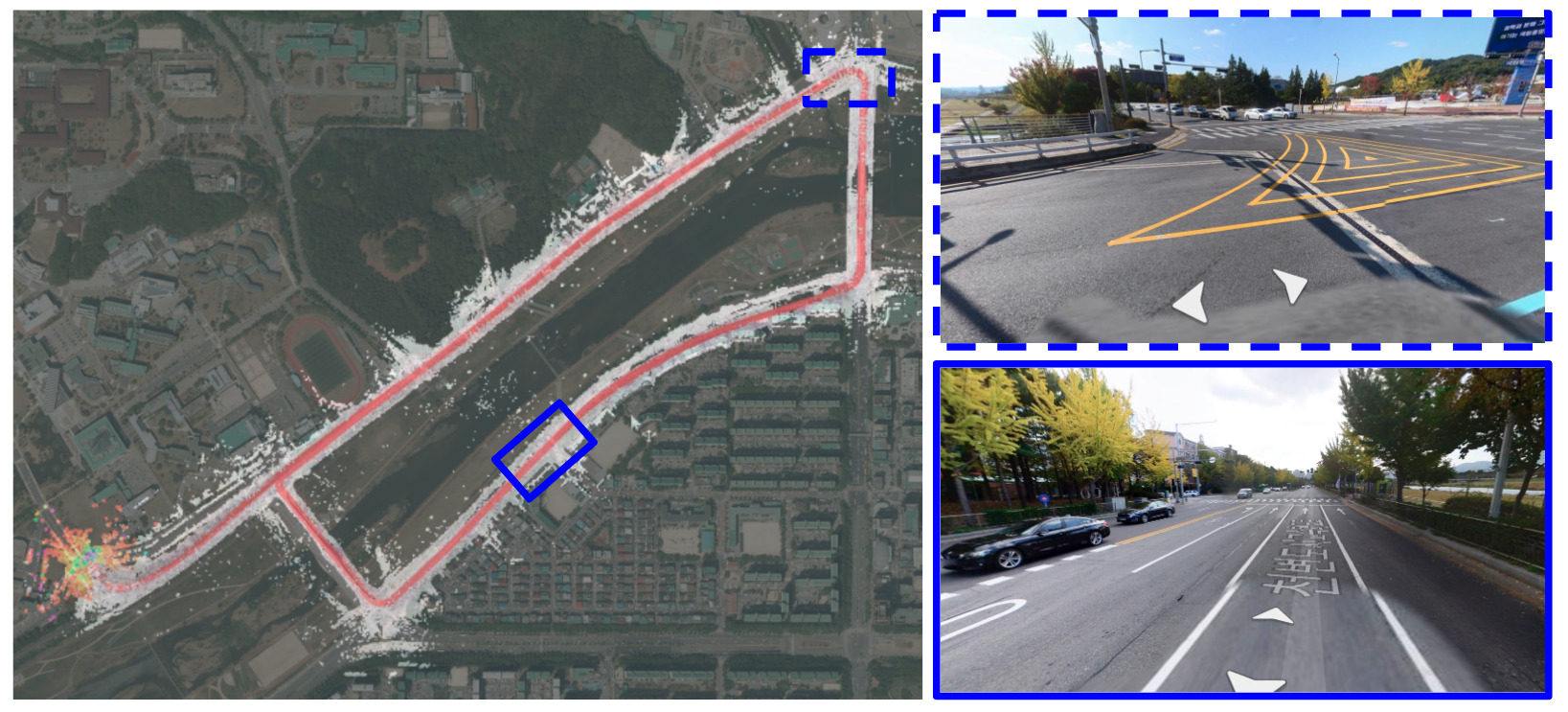}
    \label{fig:riverside1}
  } \\ 

  \subfigure[ Before-and-after trajectories of SC-LiDAR-SLAM (integrated with A-LOAM)]{%
  \includegraphics[width=0.98\columnwidth, trim = 0 0 0 0, clip]{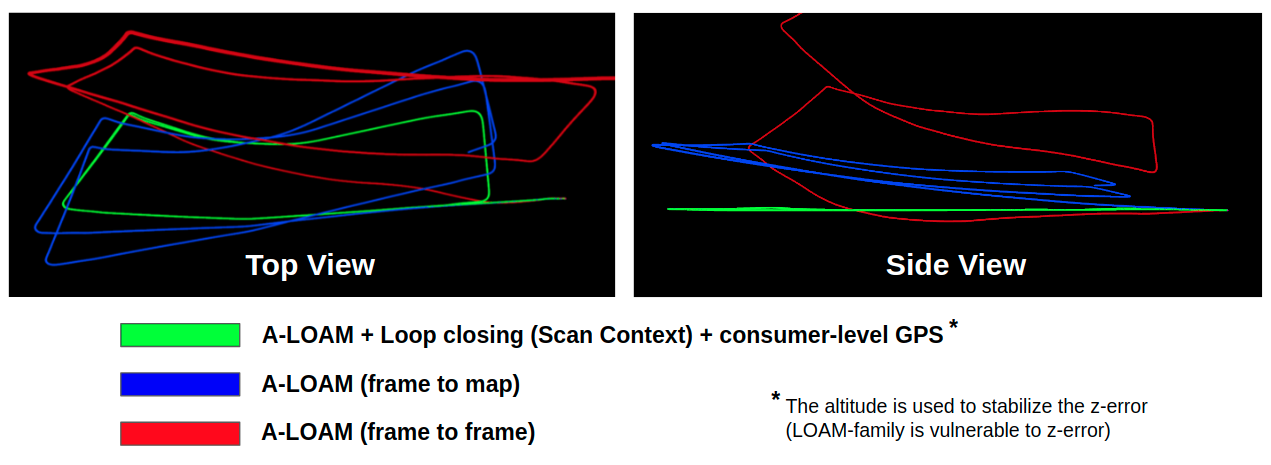}
  \label{fig:riverside2}
  }

  \subfigure[ Before-and-after point cloud maps of SC-LiDAR-SLAM (integrated with FAST-LIO)]{%
  \includegraphics[width=0.9\columnwidth, trim = 0 0 0 0, clip]{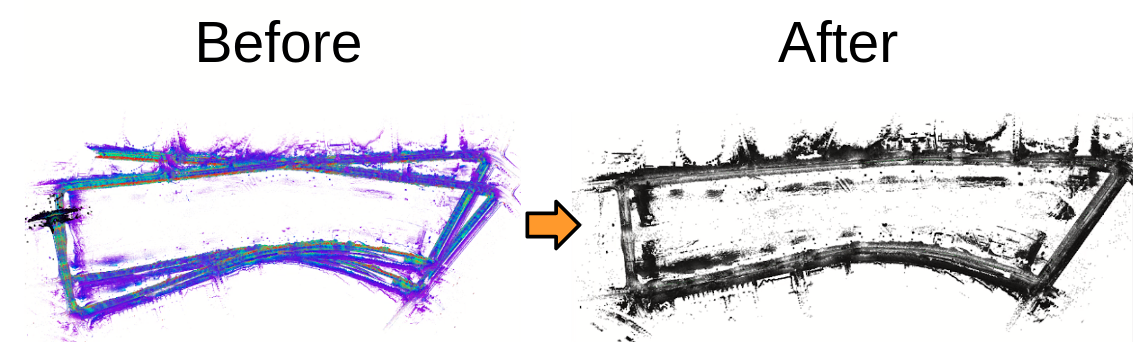}
  \label{fig:riverside3}
  } \\

  \caption{
    Tests of our robustness and versatility on a highly complex (i.e., many dynamic objects exist) urban road environment, \texttt{Riverside} sequence of \texttt{MulRan} dataset. \subref{fig:riverside2} and \subref{fig:riverside3} A-LOAM and FAST-LIO \cite{xu2021fast} are integrated, respectively. 
  }
  \label{fig:riverside}

\end{figure}


\subsection{Experimental Setup}
\label{sec:result1}

\subsubsection{Implementations}

SC-LiDAR-SLAM's main core is called SC-PGO. SC-PGO has been integrated with existing open-source LiDAR odometry methods, and the integration of the full LiDAR-SLAM systems are available via independent repositories for the ease of use: LeGO-LOAM \cite{shan2018lego}\footnote{https://github.com/irapkaist/SC-LeGO-LOAM}, LIO-SAM \cite{shan2020lio}\footnote{https://github.com/gisbi-kim/SC-LIO-SAM}, A-LOAM\footnote{https://github.com/gisbi-kim/SC-A-LOAM}, or FAST-LIO \cite{xu2021fast}\footnote{https://github.com/gisbi-kim/FAST\_LIO\_SLAM}. The implementations are in C++ and heavily relied on GTSAM \cite{dellaert2012factor} for the solver. The place recognition codes are borrowed from our previous work\footnote{https://github.com/irapkaist/scancontext}.

\subsubsection{Datasets}

Two datasets, \texttt{KITTI} datset \cite{geiger2012we} and \texttt{MulRan} dataset \cite{kim2020mulran}\footnote{https://sites.google.com/view/mulran-pr}, are selected to verify the effectiveness of SC-LiDAR-SLAM. \texttt{KITTI} dataset is a famous benchmark dataset, which is proper to compare existing methods, and \texttt{MulRan} dataset is particularly targeted to develop an algorithm that works for complex urban sites.

\subsection{Pose-graph Optimization and Point cloud Maps}
\label{sec:result2}

\textbf{KITTI dataset.} In \figref{fig:kitti05}, from the \texttt{05} sequence of \texttt{KITTI} dataset, we can see the optimized trajectory of SC-LiDAR-SLAM successfully eliminate the z-axis errors.

\textbf{MulRan dataset.} \texttt{KAIST 03} sequence's results are provided in \figref{fig:kaist} and \figref{fig:kaistmap}. The resulting point cloud map and the structures such as buildings in the map could clearly be recognized without noise from falsely registered submaps. This map correction by using the optimized loop-closed trajectory is more distinctly found in \texttt{Riverside} sequences in \figref{fig:riverside}. This environment has wide roads with multiple fast-moving cars. Particularly, there are less structured features such as buildings. The dominant unstructured things side of the road such as trees would provoke more drifts\footnote{https://youtu.be/94mC05PesvQ} of odometry methods as in the left in \figref{fig:riverside3}. SC-LiDAR- SLAM successfully resolved the odometry errors and easily integrated with various odometry methods (i.e., with A-LOAM\footnote{https://github.com/HKUST-Aerial-Robotics/A-LOAM} in \figref{fig:riverside2} and with FAST-LIO \figref{fig:riverside3}).

\subsection{Applications}
\label{sec:result3}

As mentioned in \secref{sec:method3}, SC-LiDAR-SLAM can be used for diverse useful robotic applications. In this subsection, we introduce some examples.

\subsubsection{Dynamic points removal}
\begin{figure}[!t]
  \centering
  \includegraphics[width=0.99\columnwidth, trim = 0 0 0 0, clip]{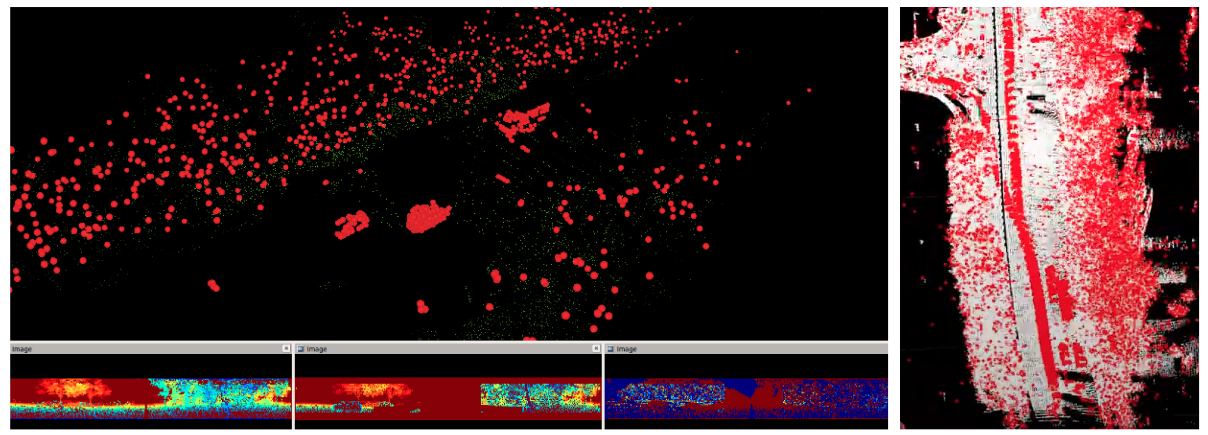}
  \caption{
    A dynamic 3D point removal example on the \texttt{Riverside} sequence. The optimized poses and saved LiDAR scans from the results in \figref{fig:riverside2} are used; those are the only ingredients for the removal without any manual inspection or training. The vivid red-colored dragged items came from a near fast-moving automobile and were well removed.
    }
  \label{fig:removert}
\end{figure}

In our other work, we proposed dynamic point removal \cite{gskim-2020-iros}\footnote{https://github.com/irapkaist/removert} only using SLAM results, not manual human intervention or labels. This framework \textit{Removert}, takes the files saved from SC-LiDAR-SLAM and remove dynamic points such as moving cars in a map as in \figref{fig:removert}.

\subsubsection{Multi-session Map Merging}
\begin{figure}[!t]
  \centering
  \includegraphics[width=0.99\columnwidth, trim = 60 20 60 20, clip]{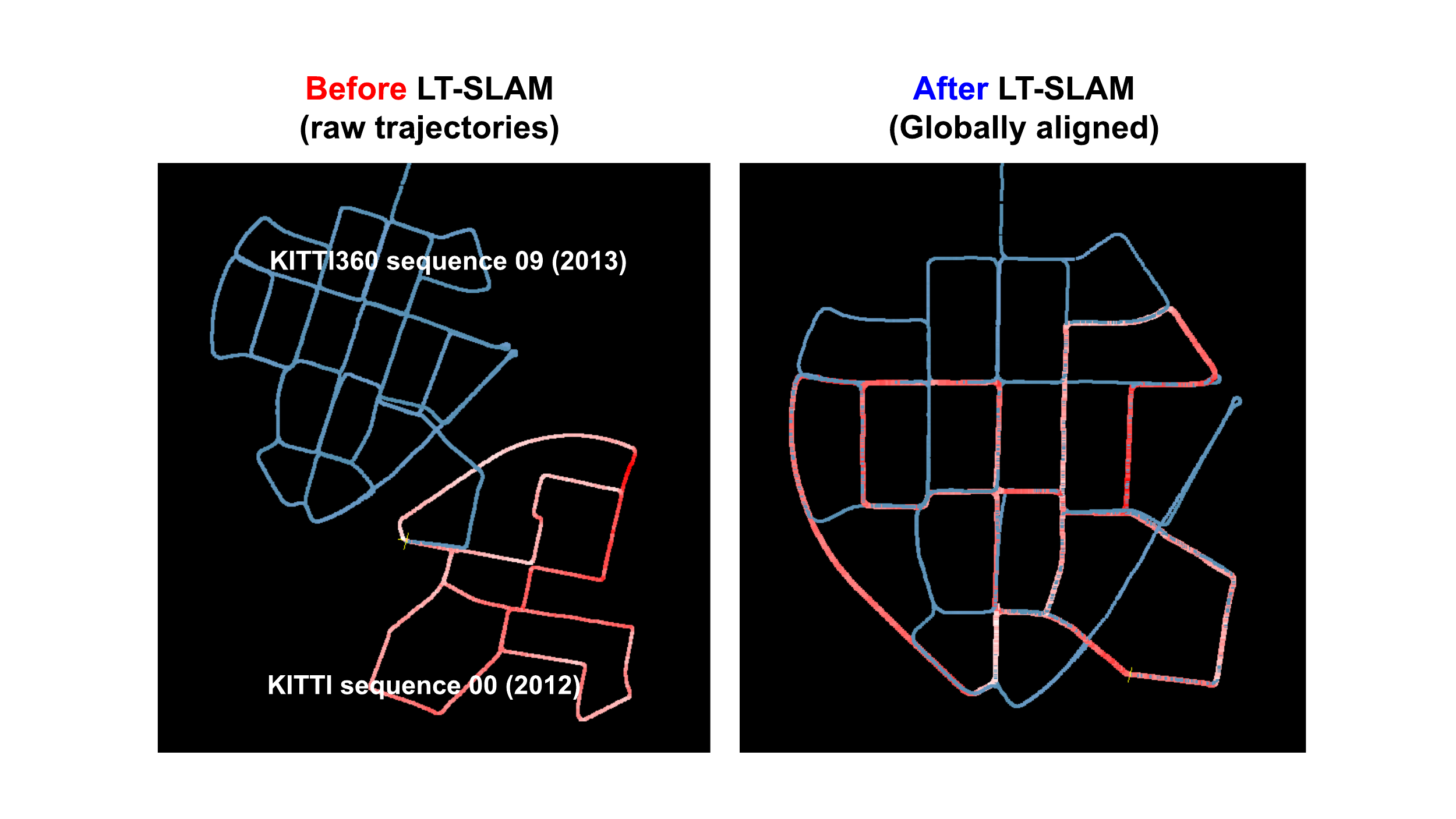}
  \caption{
    A multi-session trajectory alignment result. A set of inputs (e.g., optimized pose-graphs and the saved place data such as scans and Scan Context Descriptors) to enable this multi-session SLAM are provided by SC-LiDAR-SLAM.       
  }
  \label{fig:ltslam}
\end{figure}

In our other work \cite{kim2021lt}, we can stitch multiple LiDAR maps acquired at different timestamps by using the results of SC-LiDAR SLAM. In \figref{fig:ltslam}, the two maps (trajectories are only visualized for clearness) covered the same region are shown. They were from different datasets, \texttt{KITTI} dataset \cite{geiger2012we} and \texttt{KITTI-360} dataset \cite{Liao2021kitti360} logged at different times, 2012 and 2013 respectively. As the left of \figref{fig:ltslam}, the two maps were origianlly not aligned because their starting pose is different but they considered their own starting pose is always the zero (identity). Using the Scan Context descriptors and the initially optimized trajectories by SC-LiAR-SLAM, LT-SLAM \cite{kim2021lt}\footnote{https://github.com/gisbi-kim/lt-mapper} is capable to add inter-session loop closing factors and the two maps can be successfully aligned in a shared coordinate. This is a key for long-term map management or spatial expanding of point cloud maps from multiple distributed fleet robots.

\subsubsection{Radar SLAM}
\begin{figure}[!t]
  \centering
  \includegraphics[width=0.99\columnwidth, trim = 0 0 0 0, clip]{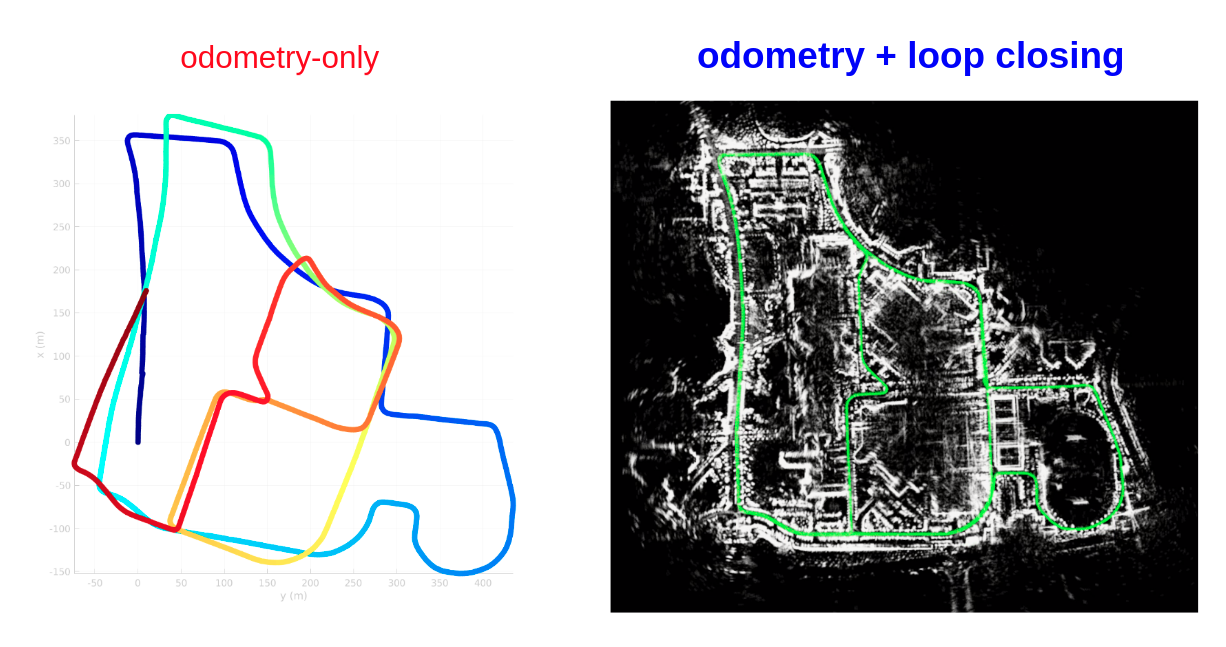}
  \caption{
    Before-and-after of SC-LiDAR-SLAM for radar sensor. In this example, Yeti radar odometry \cite{burnett_ral21} is integrated.
  }
  \label{fig:radar}
\end{figure}

SC-LiDAR-SLAM’s generic pose-graph optimization can be also applied to radar sensors. As in \figref{fig:radar}, we consider the radar data like 2D point cloud whose $z$ values are fixed. Thanks to our front-end agnostic design, the radar SLAM, and accurate radar map can be easily made without extra effort\footnote{https://github.com/gisbi-kim/navtech-radar-slam}.

\section{Conclusion}
\label{sec:conclusion}

In this paper, we proposed a front-end agnostic LiDAR SLAM system named SC-LiDAR-SLAM, and the diverse qualitative results were provided. Due to our modular design and the powerful loop closing capabilities of Scan Context\texttt{++}, we were able to easily integrate with various LiDAR odometry (or even radar odometry) methods and generate accurate point cloud maps. In future work, we would add a set of quantitative performance analyses of the proposed LiDAR SLAM system.


\renewcommand*{\bibfont}{\small}
\bibliographystyle{IEEEtranN} 
\bibliography{bib/string-short,bib/references}

\end{document}